\pdfoutput=1
\documentclass[11pt]{article}

\usepackage{EMNLP2023}

\usepackage{times}
\usepackage{latexsym}

\usepackage[T1]{fontenc}
\usepackage[utf8]{inputenc}

\newlength{\astspace}
\usepackage{microtype}
\usepackage {colortbl,array,xcolor}
\usepackage{booktabs}
\usepackage{makecell}
\usepackage{multirow}
\usepackage{autobreak}
\usepackage{lscape}
\usepackage{leipzig}
\usepackage{algorithm}
\usepackage{algpseudocode}
\usepackage{graphicx}

\let\mrm\mathrm
\usepackage{gb4e}
\let\mathrm\mrm
\noautomath

\usepackage{inconsolata}

\title{JCoLA: Japanese Corpus of Linguistic Acceptability}

\author{Taiga Someya, Yushi Sugimoto, Yohei Oseki\\
  The University of Tokyo \\
  \texttt{\{taiga98-0809, yushis, oseki\}@g.ecc.u-tokyo.ac.jp} \\
}

\begin{document}
\maketitle
\begin{abstract}
Neural language models have exhibited outstanding performance in a range of downstream tasks. However, there is limited understanding regarding the extent to which these models internalize syntactic knowledge, so that various datasets have recently been constructed to facilitate syntactic evaluation of language models across languages. In this paper, we introduce \textbf{JCoLA} (Japanese Corpus of Linguistic Acceptability), which consists of  10,020 sentences annotated with binary acceptability judgments. Specifically, those sentences are manually extracted from linguistics textbooks, handbooks and journal articles, and split into in-domain data (86 \%; relatively simple acceptability judgments extracted from textbooks and handbooks) and out-of-domain data (14 \%; theoretically significant acceptability judgments extracted from journal articles), the latter of which is categorized by 12 linguistic phenomena. We then evaluate the syntactic knowledge of 9 different types of Japanese language models on JCoLA. The results demonstrated that several models could surpass human performance for the in-domain data, while no models were able to exceed human performance for the out-of-domain data. Error analyses by linguistic phenomena further revealed that although neural language models are adept at handling local syntactic dependencies like argument structure, their performance wanes when confronted with long-distance syntactic dependencies like verbal agreement and NPI licensing.
\end{abstract}

\section{Introduction}
Neural language models, especially Transformer-based language models~\cite{Vaswani2017-my}, have exhibited outstanding performance in a range of downstream tasks~\citep{wang-etal-2018-glue, wang2019-superglue}, yet there is limited understanding regarding the extent of linguistic knowledge these models have internalized. Several studies have explored the syntactic competence of language models through acceptability judgment tasks~\citep[e.g.,][]{Linzen2016-dc, Marvin2018-na}. These and other related studies are critical as they mark the beginning of syntactic evaluations of language models, but they were limited in the scope of linguistic phenomena. In more recent times, researchers have constructed extensive datasets to facilitate more comprehensive syntactic evaluations~\cite{Warstadt2019-ru, Warstadt2020-qe, Xiang2021-mc, trotta-etal-2021-monolingual-cross, mikhailov-etal-2022-rucola}. Nonetheless, the majority of these investigations have centered around English and other European languages~\citep{Gulordava2018-yv, Warstadt2019-ru, Warstadt2020-qe, Wilcox2018-jl}, with only a handful expanding their scope to encompass non-European languages~\citep{Gulordava2018-yv, Ravfogel2018-hd}. Notably, an even smaller number of studies have addressed a broad spectrum of linguistic phenomena in languages other than English~\citep{trotta-etal-2021-monolingual-cross, Xiang2021-mc, mikhailov-etal-2022-rucola}. 

In this paper, we introduce JCoLA (Japanese Corpus of Linguistic Acceptability)~\footnote{JCoLA is available at \url{https://github.com/osekilab/JCoLA}. JCoLA is adopted as one of six tasks of JGLUE~\cite{kurihara-etal-2022-jglue}, a benchmark for natural language understanding (NLU) in Japanese.}, which consists of  10,020 sentences with acceptability judgments by linguists. Specifically, those sentences are manually extracted from linguistics textbooks, handbooks and journal articles, and split into in-domain data (86 \%; relatively simple acceptability judgments extracted from textbooks and handbooks) and out-of-domain data (14 \%; theoretically significant acceptability judgments extracted from journal articles), the latter of which is categorized by 12 linguistic phenomena. We then evaluate the syntactic knowledge of 9 different types of Japanese language models on JCoLA. The results demonstrated that several models could surpass human performance for the in-domain data, while no models were able to exceed human performance for the out-of-domain data. Error analyses by linguistic phenomena further revealed that although neural language models are adept at handling local syntactic dependencies like argument structure, their performance wanes when confronted with long-distance syntactic dependencies like verbal agreement and NPI licensing.

\begin{table*}[t]
  \centering
  \begin{tabular}{c|c|c}
    \toprule
    Language & Binary Acceptability Judgment & Minimal Pairs  \\
    \midrule 
    English & CoLA~\cite{Warstadt2019-ru} & BLiMP~\cite{Warstadt2020-qe}    \\
    \midrule 
    Italian & ItaCoLA~\cite{trotta-etal-2021-monolingual-cross} &    \\
    \midrule 
    Chinese &  & CLiMP~\cite{Xiang2021-mc}  \\
    \midrule 
    Russian & RuCoLA~\cite{mikhailov-etal-2022-rucola} & \\
    \midrule 
    Japanese & JCoLA (This work) & JBLiMP~\cite{someya-oseki-2023-jblimp} \\
    \bottomrule
  \end{tabular}
  \caption{Comparison of JCoLA and other existing datasets. As of now, there are no languages other than English for which both CoLA-style and BLiMP-style datasets are available.}
  \label{tab:relatedwork}
\end{table*}

\section{Related Work}

Acceptability judgment is a crucial aspect of human linguistic competence. It refers to the innate ability of individuals to differentiate between sentences that are grammatically correct and those that are not, even without any explicit training in grammar. For instance, when presented with two sentences, individuals can intuitively recognize which one is more acceptable or natural-sounding. Such judgments are considered the primary behavioral measure used by generative linguists to study the underlying structure of language in humans~\cite{chomskySyntacticStructures1957}. By examining acceptability judgments, linguists can gain insights into the rules that govern language and how these rules are applied by speakers of a particular language.

Historically, the evaluation of language models has been conducted using metrics such as perplexity, or based on how well the models perform on specific downstream tasks, as seen in benchmarks like GLUE~\citep{wang-etal-2018-glue}. However, in recent years, there have been efforts to assess the syntactic knowledge of language models through acceptability judgment tasks.

\citet{Linzen2016-dc} first employed minimal pairs to examine how well LSTM language models could capture subject-verb agreement in English. 
\begin{exe}
    \ex[]{The \underline{key is} on the table.}
    \ex[*]{The \underline{key are} on the table.}
\end{exe}
This and other related studies are critical as they mark the beginning of syntactic evaluations of language models. However, they were limited in the scope of linguistic phenomena considered~\citep[e.g.,][]{Marvin2018-na, Futrell2019-nz, Gulordava2018-yv}.

In light of this, more recent approaches introduced large-scale acceptability judgment corpora for targeted syntactic evaluations of language models~\cite{Warstadt2019-ru, Warstadt2020-qe}. Similar to \citet{Linzen2016-dc}, \citet{Warstadt2020-qe} constructed BLiMP (Benchmark of Linguistic Minimal Pairs) as a dataset employing minimal pairs. BLiMP  consists of 67,000 minimal pairs automatically generated across 12 types of linguistic phenomena. This enables the evaluation of language models on a wide range of linguistic phenomena, not limited to subject-verb agreement. Furthermore, similar datasets have been developed for languages other than English, allowing for comparable evaluations across various languages~\cite{Xiang2021-mc, someya-oseki-2023-jblimp}.

Concurrently, there is also an approach to targeted syntactic evaluations of language models that does not rely on minimal pairs but instead evaluates language models with binary classification tasks based on acceptability. CoLA (Corpus of Linguistic Acceptability; \citet{Warstadt2019-ru}) is the first corpus that achieves this, a dataset built by collecting sentences from syntax textbooks, handbooks, and linguistics journals. Similar datasets to CoLA have also been emerging for languages other than English~\cite{trotta-etal-2021-monolingual-cross,mikhailov-etal-2022-rucola}, though none exist for Japanese as of yet (cf. Table~\ref{tab:relatedwork}).

\section{JCoLA}
In this study, we introduce JCoLA (Japanese Corpus of Linguistic Acceptability), which will be the first large-scale acceptability judgment task dataset focusing on Japanese. JCoLA consists of sentences from textbooks and handbooks on Japanese syntax, as well as from journal articles on Japanese syntax that are published in JEAL (Journal of East Asian Linguistics), one of the prestigious journals in theoretical linguistics.

\begin{table}[t!]
\centering
\small
\begin{tabular}{lcc}
\toprule
Source & N & $\%$ \\
\midrule
\citet{gunjiJapanesePhraseStructure1987a} & 301 & 88.0 \\
\citet{inoue1976-a, inoue1976-b} & 1805 & 86.2 \\
\citet{StructureJapaneseLanguage} & 1553 & 78.0 \\
\citet{kurodaGenerativeGrammaticalStudies1965} & 332 & 91.6 \\
\citet{kurodaJapaneseSyntaxSemantics1992} & 681 & 85.5 \\
\citet{miyagawa2008} & 591 & 82.7 \\
\citet{shibatani1976} & 2209 & 83.3 \\
\citet{shibatani1990} & 387 & 90.2 \\
\citet{tsujimura1999} & 531 & 75.9 \\
\citet{tsujimura2013} & 259 & 81.1 \\
\textbf{In-Domain} & 8649 & 83.4 \\
\midrule
\citet{Abe2011-ps} & 15 & 53.3 \\
\citet{Asano2010-nw} & 92 & 63.0 \\
\citet{Bobaljik2007-pi} & 11 & 72.7 \\
\citet{Grosu2010-gl} & 11 & 18.2 \\
\citet{Grosu2012-gk} & 8 & 62.5 \\
\citet{Hayashishita2009-zo} & 34 & 76.5 \\
\citet{Ivana2007-qi} & 38 & 73.7 \\
\citet{Kishida2012-tq} & 81 & 77.8 \\
\citet{Kishimoto2008-pj} & 204 & 71.1 \\
\citet{Kishimoto2012-dg} & 90 & 61.1 \\
\citet{Miyamoto2009-jv} & 17 & 94.1 \\ 
\citet{Nishigauchi2014-cc} & 68 & 94.1 \\
\citet{Oshima2006-yg} & 25 & 96.0 \\
\citet{Saito2008-xy} & 32 & 78.1 \\
\citet{Sawada2013-cw} & 40 & 95.0 \\
\citet{Shibata2015-va} & 72 & 80.6 \\
\citet{Shimoyama2014-xl} & 51 & 92.2 \\
\citet{Sudo2015-mf} & 133 & 65.4 \\
\citet{Takahashi2006-jb} & 26 & 57.7 \\
\citet{Takahashi2010-ik} & 29 & 79.3 \\
\citet{Takano2011-qp} & 41 & 90.2 \\
\citet{Takita2009-ol} & 6 & 16.7 \\
\citet{Tenny2006-qy} & 45  & 93.3 \\
\citet{Tomioka2009-mg} & 15 & 60.0 \\
\citet{Tsujioka2011-iq} & 67 & 56.7 \\
\citet{Watanabe2010-ac} & 27 & 81.5 \\
\citet{Watanabe2013-jt} & 93 & 64.5 \\
\textbf{Out-of-Domain} & 1371 & 73.2  \\
\midrule 
\textbf{Total} & 10,020 & 82.0 \\
\bottomrule 
\end{tabular}
\caption{The number of sentences in JCoLA by source. \textit{N} is the number of sentences in a source. \textit{\%} is the percent of the acceptable sentences in a source. While \textit{In-Domain} sources are textbooks and handbooks on Japanese syntax, all the sources listed above as \textit{Out-of-Domain} are journal articles published in JEAL.}
\label{sources}
\end{table}

\subsection{Data Collection}
Sentences in JCoLA were collected from prominent textbooks and handbooks focusing on Japanese syntax. In addition to the main text, example sentences included in the footnotes were also considered for collection. We also collected acceptability judgments from journal articles on Japanese syntax published in JEAL (Journal of East Asian Linguistics): one of the prestigious journals in theoretical linguistics. Specifically, we examined all the articles published in JEAL between 2006 and 2015 (133 papers in total), and extracted 2,252 acceptability judgments from 26 papers on Japanese syntax~(Table~\ref{sources}). Acceptability judgments include sentences in appendices and footnotes, but not sentences presented for analyses of syntactic structures (e.g. sentences with brackets to show their syntactic structures).
As a result, a total of 11,984 example sentences were collected. Using this as a basis, JCoLA was constructed through the methodology explained in the following sections.

\subsection{Data Preparation}
\subsubsection{Data Preprocessing}
Among the sentences extracted through the above method, there were sentences that were not appropriate for JCoLA, a binary classification dataset based on single-sentence acceptability judgments. We either remove or modify these sentences in preprocessing. First, sentences labeled with `?', `\#', `\%', or `(?)' were removed. Additionally, sentences that did not have such labels but were noted to have variable acceptability depending on the speaker were also removed. Furthermore, duplicates, examples that were not single-sentence acceptability judgments, those containing inappropriate vocabulary, and examples whose unacceptability depends on the context were eliminated. Lastly, some sentences were found to be incomplete. In these cases, they were supplemented to form complete sentences, ensuring that the acceptability did not change. (e.g., John's book -> John's book is red.)

\subsubsection{Categorization}
A part of the data is annotated based on linguistic phenomena in order to analyze each phenomenon in detail. We categorize the 12 phenomena in JCoLA as follows~(Table~\ref{tab:minimalpair}):
\begin{table}[h]
  \centering
  \small
  \begin{tabular}{cc}
    \toprule
    Phenomenon &  \# Sentences \\
    \midrule
    \textsc{Argument structure} & 545 \\
    \textsc{Filler-gap} & 257 \\
    \textsc{Morphology} & 159 \\   
    \textsc{Nominal structure} & 150 \\
    \textsc{Quantifier} & 127 \\ 
    \textsc{Verbal agreement} & 105 \\
    \textsc{Binding} & 101 \\
    \textsc{Ellipsis} & 44 \\ 
    \textsc{Island effects} & 19 \\
    \textsc{NPI/NCI} & 12 \\
    \textsc{Control/raising} & 11 \\
    \textsc{Simple} & 71 \\
    \bottomrule
  \end{tabular}
  \caption{Number of sentences by phenomenon in out-of-domain data. Note that the examples in JCoLA could be categorized into multiple phenomena. }
  \label{tab:minimalpair}
\end{table}

\noindent {\textbf{Argument Structure:}} acceptability judgements
based on the order of arguments (\ref{oa}) and case marking (\ref{cm}).

\begin{exe}
    \ex 
    \begin{xlist}
        \ex[]{
        \gll Ken-ni tegami-ga todoita.\\
            Ken-{\textsc{dat}} letter-{\textsc{nom}} reached\\
        \glt `A letter reached Ken.'}\label{oa}   %Takano 2011, (1b)
        \ex[*]{
        \gll Taroo-ga Hanako-o au.\\
            Taroo-{\textsc{nom}} Hanako-{\textsc{acc}} see\\
            \glt `*Taroo sees Hanako'}\label{cm}%Takahashi 2006
    \end{xlist}
\end{exe} 

\noindent {\textbf{Binding:}} acceptability judgements based on the
binding of noun phrases. For instance, this includes
reflexive binding (\ref{rb}) and the coreference resolution of anaphors (\ref{cra}).

\begin{exe}
    \ex 
    \begin{xlist}
    \ex[]{
        \gll Ken-ga zibun-no heya-ni modotta\\
            Ken-{\textsc{nom}} self-{\textsc{gen}} room-{\textsc{dat}} returned\\
        \glt `Ken returned to his room.'}\label{rb} %Takano 2011, 15a
    \ex[?*]{
    \gll Hazimete soitu-ni au hito-ga kenasu no-wa dare-o desu ka?\\
     for-the-first-time him-{\textsc{dat}} see person-{\textsc{nom}} criticize that-{\textsc{top}} who-{\textsc{acc}} is Q\\ 
    \glt `?*Who is it that people who see him for the first time criticize?'}\label{cra}  %Takahashi 2006, 16a
    \end{xlist}
\end{exe}

\noindent {\textbf{Control/Raising:}} acceptability judgements
based on predicates that are categorized as control
or raising.

\begin{exe}
        \ex\label{control}
            \gll John-wa ie-o tukuri-sokoneta\\
                 John-{\textsc{top}} house-{\textsc{acc}} make-to-fail-{\textsc{past}}\\
            \glt `John failed to make a house.'%Kishita-Sato 2012, 19aii
\end{exe}

\noindent {\textbf{Ellipsis:}} acceptability judgements based on the
possibility of omitting elements in the sentences.
For instance, this includes nominal (\ref{ne}) and adjunct ellipsis (\ref{ae}).

\begin{exe}
    \ex 
    \begin{xlist}
        \ex[]{
        \gll Taroo-ga zibun-o hihansita-ra Hanako-wa hometa.\\
             Taroo-{\textsc{nom}} self-{\textsc{acc}} criticized-when Hanako-{\textsc{top}} praised\\
       \glt `When Taroo criticized himself, Hanako praised.'}\label{ne} % Takahashi 2006, 29q

        \ex[*]{
         \gll Taroo-ga sono riyuu de kaikosareta atode, Hanako-mo kaikosareta.\\
              Taroo-{\textsc{nom}} that reason for was-fired after Hanako-also was-fired\\
        \glt `*After Taroo was fired for that reason, Hanako was fired too.'}\label{ae}%Takahashi 2006, 59
        \end{xlist}
\end{exe}

\noindent {\textbf{Filler-gap:}} acceptability judgements based on
the dependency between the moved element and
the gap. For instance, this includes comparatives (\ref{cp}) and cleft sentences (\ref{cl}).

\begin{exe}
    \ex 
    \begin{xlist}
        \ex\label{cp}
        \gll Mary-wa John-ga kaita yori nagai ronbun-o kaita.\\
             Mary-{\textsc{top}} John-{\textsc{nom}} wrote than long paper-{\textsc{acc}} wrote\\
        \glt   `Mary wrote a longer paper than John wrote'
        \ex\label{cl}
        \gll Taroo-ga atta no-wa Hanako-ni da.\\
             Taroo-{\textsc{nom}} saw that-{\textsc{top}} Hanako-{\textsc{dat}} is\\
         \glt `It was Hanako that Taroo saw.' %Takahashi 2006, (10a)
    \end{xlist}
\end{exe}

\noindent {\textbf{Island Effects:}} acceptability judgements based
on the restrictions on filler-gap dependencies such
as wh-movements.

\begin{exe}
    \ex[*]{
    \gll Taroo-wa Hanako-ga naze kare-no tegami-o suteta kara okotteiru no?\\
         Taroo-{\textsc{top}} Hanako-{\textsc{nom}} why he-{\textsc{gen}} letter-{\textsc{acc}} discarded because be.angry C\\
    \glt `*Why is Taro angry because Hanako discarded his letter?'}%Takita 2009, (27a)
\end{exe}

\noindent {\textbf{Morphology:}} acceptability judgements based on
the morphology. For instance, it includes idioms.

\begin{exe}
    \ex 
    \gll Taroo-no kotoba-wa hi-ni abura-o sosoida.\\
        Taroo-{\textsc{gen}} words-{\textsc{top}} fire-{\textsc{dat}} oil-{\textsc{acc}} pour\\
    \glt `Taroo's words made the situation worse' %Kishimoto 2008, (9)
\end{exe}

\noindent {\textbf{Nominal Structure:}} acceptability judgements
based on the internal structure of noun phrases.

\begin{exe}
    \ex 
    \gll  amen-no hi-wa kiraida\\
          rainy day-{\textsc{top}} hate.be\\
    \glt `I hate rainy days.' %Saito et al. 2008, (27)
\end{exe}

\noindent {\textbf{NPI/NCI:}} acceptability judgements based
on the restrictions on where negative polarity/concord items
(NPIs/NCIs) can appear. For instance, NCIs include {\textit{daremo}}.

\begin{exe}
    \ex 
    \gll Daremo monku-o iw-anakat-ta.\\
         who-{\textsc{mo}} complaint-{\textsc{acc}} say-{\textsc{neg}}-{\textsc{past}}\\
    \glt `Nobody complained.'   %Shibatani 2015, 28b
\end{exe}

\noindent {\textbf{Quantifier:}} acceptability judgements based on
the distribution of quantifiers such as floating quantifiers.

\begin{exe}
    \ex 
    \gll John-wa hon-o san-satsu katta.\\
         John-{\textsc{top}} book-{\textsc{acc}} three-{\textsc{cl}} bought\\
    \glt `John bought three books.' %Watanabe 2010, (23c)
\end{exe}

\noindent {\textbf{Verbal Agreement:}} acceptability judgements based on the dependency between subjects and verbs. Japanese doesn’t have the same kind of subject-verb agreement as in English. Instead, this includes the linguistic phenomena such as subject honorification where the social status of subjects are reflected in the morphology of verbs.

\begin{exe}
    \ex 
    \begin{xlist}
        \ex[]{
        \gll Ito-sensei-ga Mary-o o-home-ni-nat-ta.\\ 
             Ito-teacher-{\textsc{nom}} Mary-{\textsc{acc}} {\textsc{hon}}-praise-{\textsc{lv}}-{\textsc{past}}\\
        \glt `Prof. Ito praised Mary.'}
        \ex[*]{
        \gll Mary-ga Ito-sensei-o o-home-ni-nat-ta.\\
              Mary-{\textsc{nom}} Ito-teacher-{\textsc{acc}} {\textsc{hon}}-praise-{\textsc{lv}}-{\textsc{past}}\\ %LV: light verb
        \glt `Mary praised Prof. Ito.'} %Kishimoto 2012, (3)a,b
    \end{xlist}
\end{exe}

\noindent {\textbf{Simple: }} 
acceptability judgements that do not have marked syntactic structures. For instance, it includes a simple transitive sentence. 

\begin{exe}
    \ex 
    \gll John-ga hon-o yonda\\
         John-{\textsc{nom}} book-{\textsc{acc}} read\\
    \glt `John read a book.' %Sudo 2015, (26a) 
\end{exe}

Sentences that do not fall into these 12 phenomena were deleted.

Note that the examples in JCoLA could be categorized in multiple phenomena. For example, the following sentence includes a classifier {\textit{mit-tu}} `three', which is a quantifier-binder and a variable {\textit{soko}} `it', which gets a bound variable interpretation. Thus, this is a combination of binding and quantifier phenomena. 

\begin{exe}
    \ex 
    \gll Mit-tu-izyoo-no kaisya-o soko-no syain-ga hihansi-ta\\
         3-{\textsc{cl}}-or.more-{\textsc{gen}} company-{\textsc{acc}} it-{\textsc{gen}} employee-{\textsc{nom}} criticized-{\textsc{past}}\\
    \glt `Three companies, its employee(s) criticized.'
\end{exe}

\subsection{Data Validation}
As a reference for the upper limit of accuracy in JCoLA, human acceptability judgment experiments were conducted on Lancers\footnote{\url{https://www.lancers.jp/}} with a subset of the JCoLA data. Specifically, we conducted acceptability judgment experiments on 200 sentences sampled from the in-domain data and all the sentences in the out-of-domain data, making a total of 1,951 sentences. To reduce the burden on each annotator, the sentences were divided into 38 groups of 50 sentences and one group of 51 sentences. Each annotator performed a forced-choice binary acceptability judgment task on 50 or 51 sentences. For the out-of-domain data, if the results of the acceptability judgment experiment did not match between the human majority vote and the JCoLA annotation, that data was removed. As a result, 380 instances were deleted, leaving 1,371 instances in the out-of-domain data. 
The results showed that for the in-domain data, the individual agreement with JCoLA was $75.9\%$, and the majority vote agreement with JCoLA was $79.5\%$. For the out-of-domain data, the individual agreement with JCoLA was $85.4\%$, and the majority vote agreement with JCoLA was $100.0\%$ (due to the aforementioned data removal).

\subsection{Data Split}
While CoLA includes out-of-domain data in addition to the standard train/dev/test splits to assess whether overfitting occurs to specific sources or linguistic phenomena within the training data, JCoLA will also incorporate out-of-domain data. However, in JCoLA, the data collected from journal articles in JEAL are designated as out-of-domain. This is because JCoLA aims to evaluate whether language models can generalize to more complex linguistic phenomena (cf.~Class I\hspace{-.1em}I\hspace{-.1em}I judgement, see \citealt{Marantz2005-dx}; \citealt{Linzen2018-oa}) after learning relatively simple grammatical rules (Class I\hspace{-.1em}I judgement). The in-domain data is split into training data (6,919 instances), development data (865 instances), and test data (865 instances). On the other hand, the out-of-domain data is only used for evaluation, and divided into development data (685 instances) and test data (686 instances).

\section{Experiments}
\subsection{Models}
In this paper, we evaluate some pretrained Japanese and multilingual neural language models on JCoLA.
Specifically, we evaluate nine different neural language models provided by different organizations, which are different in size, method of morphological analysis and tokenization, and training corpus.

\paragraph{BERT}
We evaluate three different types of BERT language models provided by Tohoku University NLP group\footnote{\url{https://github.com/cl-tohoku}}: Tohoku $\text{BERT}_{\text{BASE}}$\footnote{\url{https://huggingface.co/cl-tohoku/bert-base-japanese-v2}}, Tohoku $\text{BERT-char}_{\text{BASE}}$\footnote{\url{https://huggingface.co/cl-tohoku/bert-base-japanese-char-v2}} and Tohoku $\text{BERT}_{\text{LARGE}}$\footnote{\url{https://huggingface.co/cl-tohoku/bert-large-japanese}}. These models are trained on the Japanese version of Wikipedia. The texts are first tokenized by MeCab~\citep{kudo-etal-2004-applying} and then split into subwords by BPE~\citep{sennrich-etal-2016-neural}.\footnote{For Tohoku $\text{BERT-char}_{\text{BASE}}$, the texts are segmented into characters.} Tohoku $\text{BERT}_{\text{BASE}}$ and Tohoku $\text{BERT-char}_{\text{BASE}}$ have 12 layers, 12 attention heads, and 768-dimensional hidden states, while Tohoku $\text{BERT}_{\text{LARGE}}$ has 24 layers, 16 attention heads, and 1024-dimensional hidden states. 

In addition, we evaluate a BERT language model provided by NICT~(NICT $\text{BERT}_{\text{BASE}}$).\footnote{\url{https://direct.nict.go.jp/}}
The model configuration is the same as Tohoku $\text{BERT}_{\text{BASE}}$ and Tohoku $\text{BERT-char}_{\text{BASE}}$.
\paragraph{Japanese RoBERTa}
We also evaluate three variants of RoBERTa language models provided by Kawahara Lab. at Waseda University\footnote{\url{https://nlp-waseda.jp/en/}}: Waseda $\text{RoBERTa}_\text{BASE}$\footnote{\url{https://huggingface.co/nlp-waseda/roberta-base-japanese}}, Waseda $\text{RoBERTa-seq128}_\text{LARGE}$\footnote{\url{https://huggingface.co/nlp-waseda/roberta-large-japanese}} and Waseda $\text{RoBERTa-seq512}_\text{LARGE}$\footnote{\url{https://huggingface.co/nlp-waseda/roberta-large-japanese-seq512}}. These models are trained on the Japanese version of Wikipedia and the Japanese portion of CC-100. The texts are first tokenized by Juman++~\citep{morita-etal-2015-morphological} and then split into subwords using Sentence Piece~\cite{kudo-richardson-2018-sentencepiece} with a unigram language model~\cite{kudo-2018-subword}. Waseda $\text{RoBERTa}_\text{BASE}$ has 12 layers, 12 attention heads, and 768-dimensional hidden states. Waseda $\text{RoBERTa-seq128}_\text{LARGE}$ and Waseda $\text{RoBERTa-seq512}_\text{LARGE}$ both have 24 layers, 16 attention heads, and 1024-dimensional hidden states, but are trained with the maximum sequence length of 128 and 512, respectively.

\paragraph{XLM-RoBERTa}
To compare the performance of monolingual and multilingual language models on JCoLA, we also evaluate two multilingual language models with different parameter sizes: XLM-$\text{RoBERTa}_\text{BASE}$\footnote{\url{https://huggingface.co/xlm-roberta-base}} and XLM-$\text{RoBERTa}_\text{LARGE}$\footnote{\url{https://huggingface.co/xlm-roberta-large}}. These models are trained on multilingual Common Crawl~\cite{wenzek-etal-2020-ccnet} and the train texts are directly tokenized using Sentence Piece~\cite{kudo-richardson-2018-sentencepiece} with a unigram language model~\cite{kudo-2018-subword}. XLM-$\text{RoBERTa}_\text{BASE}$ has 12 layers, 12 attention heads and 768-dimensional hidden states. XLM-$\text{RoBERTa}_\text{LARGE}$ has 24 layers, 16 attention heads and 1024-dimensional hidden states.

\subsection{Training Settings}
Each language model is trained for five epochs with AdamW optimizer~\cite{loshchilov_decoupled_2019} and linear warmup with a warmup ratio of 0.1. In addition, the language models are trained using three different learning rates (5e-5, 3e-5, and 2e-5) and we evaluate models which achieved the highest Matthews Correlation Coefficient (MCC; ~\citet{MATTHEWS1975442}) on the development data. This evaluation metric is an evaluation metric suitable for unbalanced binary classifiers also used in \citet{Warstadt2019-ru}. For each configuration, we trained 20 models with different random seeds to mitigate the effect of randomness. The score for each language model is calculated as the average across 20 different random seeds, but we ignore those results where the models achieved less than zero MCC score on the development set, as in \citet{warstadt2020linguistic}.
\begin{table*}[t]
\centering
\begin{tabular}{ccccc}
\toprule
                      \multirow{2}{*}{Model} & \multicolumn{2}{c}{In-domain} & \multicolumn{2}{c}{Out-of-domain} \\
                      & Acc. & MCC & Acc. & MCC \\
\midrule
           Tohoku BERT base & 0.838 ± 0.007 &  0.350 ± 0.027 &          0.753 ± 0.007 &          0.247 ± 0.028 \\
    Tohoku BERT base (char) & 0.815 ± 0.007 & 0.236 ± 0.032 &           0.740 ± 0.008 &          0.164 ± 0.057 \\
          Tohoku BERT large & 0.835 ± 0.004 & 0.346 ± 0.022 &          0.769 ± 0.008 &          0.309 ± 0.033 \\
             NICT BERT base & 0.841 ± 0.007 &  0.360 ± 0.036 &          0.773 ± 0.006 &          0.329 ± 0.023 \\
        Waseda RoBERTa base & 0.855 ± 0.008 & 0.404 ± 0.037 &          0.781 ± 0.017 &          0.355 ± 0.069 \\
Waseda RoBERTa large (s128) & \textbf{0.864 ± 0.007} & \textbf{0.461 ± 0.032} &         \textbf{0.822 ± 0.012} &          \textbf{0.507 ± 0.038} \\
Waseda RoBERTa large (s512) & 0.860 ± 0.009 & 0.419 ± 0.054 &          0.810 ± 0.010 &          0.465 ± 0.032 \\
           XLM RoBERTa base & 0.827 ± 0.004 & 0.172 ± 0.055 &          0.745 ± 0.009 &          0.176 ± 0.063 \\
          XLM RoBERTa large & 0.831 ± 0.007 & 0.214 ± 0.128 &          0.772 ± 0.008 &           0.320 ± 0.033 \\
\midrule 
Human (Individual) & 0.760 & 0.384 &  0.854 & 0.653\\
Human (Majority vote) & 0.795 & 0.437 & 1.000 & 1.000 \\
\bottomrule
\end{tabular}
\caption{Performance of each language model on JCoLA out-of-domain test set. The score for each language model is calculated as the average across 20 different random seeds, but we ignore those results where the models achieved less than zero MCC score on the development set, as in \citet{warstadt2020linguistic}. The best performance across models is indicated in bold.}
\label{tab:overall}
\end{table*}

\section{Results and Discussion}
\subsection{Overall performance}

Table~\ref{tab:overall} presents the Matthews Correlation Coefficient (MCC) and accuracy of various models on the in-domain and out-of-domain data, along with human performance. In the in-domain data, several models demonstrate performance surpassing that of human individuals. However, in the case of out-of-domain data, none of the models were able to exceed human performance. This suggests that the language models may not necessarily capture the complex linguistic phenomena addressed in theoretical linguistics (Class \mbox{I\hspace{-.1em}I\hspace{-.1em}I} judgement).
However, while the majority of models have lower performance on out-of-domain data compared to in-domain data, some models perform better on out-of-domain data. These models appear to be generalizing the linguistic phenomena observed in in-domain data correctly and are somewhat able to judge acceptability even for more complex linguistic phenomena.\footnote{Interestingly, the models that exhibited higher performance on out-of-domain data all utilized Sentence Piece with a unigram language model for tokenization, indicating the possibility that this choice of tokenization method may have contributed in some way to their performance.}

\subsection{Performance by phenomenon}

\begin{figure*}[t]
    \centering
    \includegraphics[scale=0.415]{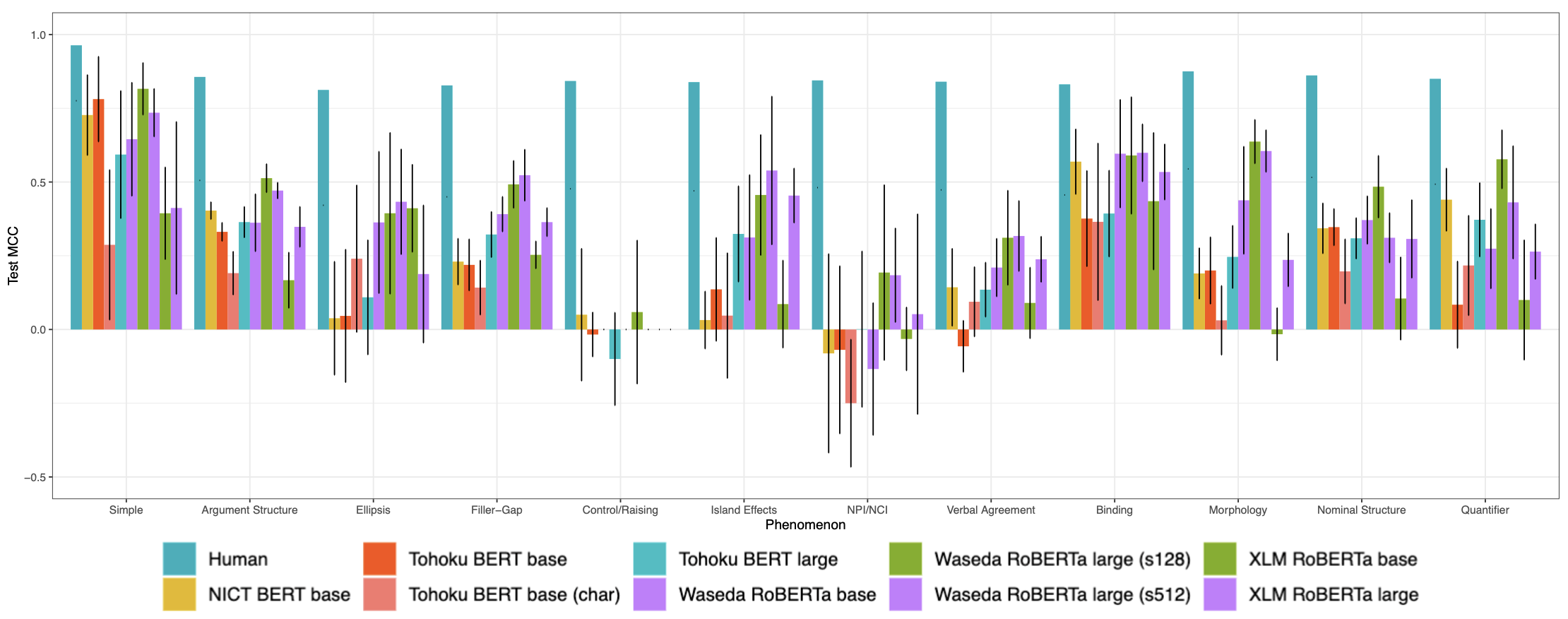}
    \caption{Performance of each language model on JCoLA out-of-domain test set by phenomenon. The MCC score for each language model is calculated as the average across 20 different random seeds, but we ignore those results where the models achieved less than zero MCC score on the development set, as in \citet{warstadt2020linguistic}. Error bars mark the mean $\pm1$ SD.}
    \label{fig:by-phenomenon}
\end{figure*}

Figure~\ref{fig:by-phenomenon} shows the Matthews Correlation Coefficient (MCC) values for each linguistic phenomenon in the out-of-domain test set across different models. Notably, almost all models demonstrate high accuracy in the Simple category, which suggests that they are capable of accurately capturing this linguistic phenomenon, even with sentences from sources not seen during training. However, for other phenomena, the performance is generally lower than that for Simple. In fact, the average MCC across linguistic phenomena, excluding Simple, is 0.248, which is significantly lower than the 0.599 observed for Simple. This suggests that while language models can effectively learn relatively simple linguistic phenomena (Class I\hspace{-.1em}I judgement) as presented in textbooks and handbooks of syntactic theory, they may not necessarily be able to generalize to more complex linguistic phenomena (Class \mbox{I\hspace{-.1em}I\hspace{-.1em}I} judgement).

Furthermore, upon examining the performance of language models on different phenomena, it becomes apparent that language models perform relatively well on certain linguistic phenomena, such as binding, argument structure, and filler-gap, but struggle with others. Relatively high performance in Binding could be attributed to the fact that the proportion of positive examples for Binding is $93.1\%$, significantly higher than the overall $73.2\%$ for the out-of-domain data. For Argument Structure, many sentences only require capturing relatively local dependencies related to the order of arguments and/or case marking, such as in the example below. 

\begin{exe}
    \ex 
    \gll John-ga hon-o/*-ni yonda\\
         John-{\textsc{nom}} book-{\textsc{acc}/*\textsc{dat}} read\\
    \glt `John read a book.' %Sudo 2015, (26a) 
\end{exe}

Regarding filler-gap, even though it generally involves complex linguistic phenomena such as wh-movement, the presence of a relatively large number of sentences involving simpler comparison phenomena could be contributing to the higher accuracy.

\begin{exe}
\ex
        \gll Mary-wa John-ga kaita yori nagai ronbun-o kaita.\\
             Mary-{\textsc{top}} John-{\textsc{nom}} wrote than long paper-{\textsc{acc}} wrote\\
        \glt   `Mary wrote a longer paper than John wrote'
\end{exe}
On the other hand, language models show lower accuracy on linguistic phenomena such as NPI/NCI and verbal agreement. This could be because NPI/NCI and verbal agreement often require capturing relatively long-distance dependencies, as seen in the examples below.\footnote{The results for control/raising were not considered to be reliable due to the small sample size, and they were excluded from the analysis.}

\begin{exe}
    \ex 
    \begin{xlist}
        \ex[]{
        \gll \textbf{Ito-sensei-ga} Mary-o \textbf{o}-home-\textbf{ni-nat}-ta.\\ 
             Ito-teacher-{\textsc{nom}} Mary-{\textsc{acc}} {\textsc{hon}}-praise-{\textsc{lv}}-{\textsc{past}}\\
        \glt `Prof. Ito praised Mary.'}
        \ex[*]{
        \gll \textbf{Mary-ga} Ito-sensei-o \textbf{o}-home-\textbf{ni-nat}-ta.\\
              Mary-{\textsc{nom}} Ito-teacher-{\textsc{acc}} {\textsc{hon}}-praise-{\textsc{lv}}-{\textsc{past}}\\
        \glt `Mary praised Prof. Ito.'} %Kishimoto 2012, (3)a,b
    \end{xlist}
\end{exe}

\begin{exe}
    \ex 
    \gll \textbf{Daremo} monku-o iw-\textbf{anakat}-ta.\\
         who-{\textsc{mo}} complaint-{\textsc{acc}} say-{\textsc{neg}}-{\textsc{past}}\\
    \glt `Nobody complained.'   %Shibatani 2015, 28b
\end{exe}

Overall, the analysis by linguistic phenomenon highlights the strengths and limitations of language models in capturing various linguistic phenomena. While they are adept at handling simpler structures, their performance wanes when confronted with more complex linguistic phenomena, especially those requiring long-distance dependencies.

\section{Conclusion}
In this paper, we introduced JCoLA (Japanese Corpus of Linguistic Acceptability), which consists of  10,020 sentences annotated with binary acceptability judgments. Specifically, those sentences were manually extracted from linguistics textbooks, handbooks and journal articles, and split into in-domain data (86 \%; relatively simple acceptability judgments extracted from textbooks and handbooks) and out-of-domain data (14 \%; theoretically significant acceptability judgments extracted from linguistics journals), the latter of which was categorized by 12 linguistic phenomena. We then evaluated the syntactic knowledge of 9 different types of Japanese language models on JCoLA. The results demonstrated that several models could surpass human performance for the in-domain data, while no models were able to exceed human performance for the out-of-domain data. Error analyses by linguistic phenomena further revealed that although neural language models are adept at handling local syntactic dependencies like argument structure, their performance wanes when confronted with long-distance syntactic dependencies like verbal agreement and NPI licensing.

\section*{Limitations}
All the sentences included in JCoLA have been extracted from textbooks, handbooks and journal articles on theoretical syntax. Therefore, those sentences are guaranteed to be theoretically meaningful, making JCoLA a challenging dataset. However, the distribution of linguistic phenomena directly reflects that of the source literature and thus turns out to be extremely skewed. Indeed, as can be seen in Table~\ref{tab:minimalpair}, while the number of sentences exceeds 100 for most linguistic phenomena, there are several linguistic phenomena for which there are only about 10 sentences. In addition, since it is difficult to force language models to interpret sentences given specific contexts, those sentences whose unacceptability depends on contexts were inevitably removed from JCoLA. This removal process resulted in the deletion of unacceptable sentences from some linguistic phenomena (such as ellipsis), consequently skewing the balance between acceptable and unacceptable sentences (with a higher proportion of acceptable sentences).

\section*{Acknowledgements}
This work was supported by JST PRESTO Grant Number JPMJPR21C2, Japan. 

\bibliography{anthology,custom}
\bibliographystyle{acl_natbib}

\end{document}